\documentclass[letterpaper]{article} % DO NOT CHANGE THIS
\usepackage{aaai25}  % DO NOT CHANGE THIS
\usepackage{times}  % DO NOT CHANGE THIS
\usepackage{helvet}  % DO NOT CHANGE THIS
\usepackage{courier}  % DO NOT CHANGE THIS
\usepackage[hyphens]{url}  % DO NOT CHANGE THIS
\usepackage{graphicx} % DO NOT CHANGE THIS
\usepackage{natbib}  % DO NOT CHANGE THIS AND DO NOT ADD ANY OPTIONS TO IT
\usepackage{caption} % DO NOT CHANGE THIS AND DO NOT ADD ANY OPTIONS TO IT
\usepackage{algorithm}
\usepackage{algpseudocode}

\usepackage{newfloat}
\usepackage{listings}
\DeclareCaptionStyle{ruled}{labelfont=normalfont,labelsep=colon,strut=off} % DO NOT CHANGE THIS
\floatstyle{ruled}
\newfloat{listing}{tb}{lst}{}
\floatname{listing}{Listing}
\usepackage{subcaption}
\usepackage{algorithm}
\usepackage{amsmath}
\usepackage{amssymb}
\usepackage{booktabs}
\usepackage{multirow}

\title{FedSPU: Personalized Federated Learning for Resource-Constrained Devices with Stochastic Parameter Update}
\author {
    Ziru Niu\textsuperscript{\rm 1},
    Hai Dong\textsuperscript{\rm 1}\thanks{Corresponding author.},
    A. K. Qin\textsuperscript{\rm 2}
}
\affiliations {
    \textsuperscript{\rm 1}RMIT University, Melbourne, VIC 3000, Australia\\
    \textsuperscript{\rm 2}Swinburne University of Technology, Hawthorn, VIC 3122, Australia\\
    ziru.niu@student.rmit.edu.au, hai.dong@rmit.edu.au, 
    kqin@swin.edu.au
}
\usepackage{bibentry}
\begin{document}

\maketitle

\begin{abstract}
Personalized Federated Learning (PFL) is widely employed in the Internet of Things (IoT) to handle high-volume, non-iid client data while ensuring data privacy. However, heterogeneous edge devices owned by clients may impose varying degrees of resource constraints, causing computation and communication bottlenecks for PFL. Federated Dropout has emerged as a popular strategy to address this challenge, wherein only a subset of the global model, i.e. a \textit{sub-model}, is trained on a client's device, thereby reducing computation and communication overheads. Nevertheless, the dropout-based model-pruning strategy may introduce bias, particularly towards non-iid local data. When biased sub-models absorb highly divergent parameters from other clients, performance degradation becomes inevitable. In response, we propose federated learning with stochastic parameter update (FedSPU). Unlike dropout that tailors local models to small-size sub-models, FedSPU maintains the full model architecture on each device but randomly freezes a certain percentage of neurons in the local model during training while updating the remaining neurons. This approach ensures that a portion of the local model remains personalized, thereby enhancing the model's robustness against biased parameters from other clients. Experimental results demonstrate that FedSPU outperforms federated dropout by 4.45\% on average in terms of accuracy. Furthermore, an introduced early stopping scheme leads to a significant reduction of the training time in FedSPU by \(25\%\sim71\%\) while maintaining high accuracy. The experiment code is available at: https://github.com/ZiruNiu0/FedSPU.
\end{abstract}

%\begin{links}
%     \link{Code}{https://github.com/ZiruNiu0/FedSPU}
%\end{links}

\section{Introduction}

Federated Learning (FL) is a distributed machine learning paradigm that allows edge devices to collaboratively train a model without revealing private data \cite{fedavg}. However, the efficacy of FL in real-world IoT systems is usually impeded by both data and system heterogeneities of IoT clients. First, clients comprise edge devices from various geographical locations collecting data \emph{that are naturally non-independent identical (non-iid)}. In such a scenario, a single global model struggles to generalize across all local datasets \cite{pflsurvey, hermes}. Second, IoT clients consist of physical devices with varying processor, memory, and bandwidth capabilities \cite{mocha, flRCsurvey}. Among these devices, some resource-constrained devices might be incapable of training the entire global model with a too complex structure.\par

To overcome the non-iid data problem, the Personalized Federated Learning (PFL) framework is introduced by empowering each client to maintain a unique local model tailored to its local data distribution. To address system heterogeneity, the technique of \emph{federated dropout} (i.e. model pruning) is employed. Resource-constrained devices are allowed to train a \textit{sub-model}, which is a subset of the global model. This approach reduces computation and communication overheads for training and transmitting the sub-model, aiding resource-constrained devices in overcoming computation and communication bottlenecks \cite{randdrop, fjord}.
\par 
Various PFL frameworks have been developed to tackle the non-iid data challenge \cite{adaptiveFT, pflmeta, pflayer, clusterfl, mocha, pflod, fedem, fedtha, pfedme, threepfl, apfl, fedala}.
Nevertheless, the inherent communication or computation bottleneck of resource-constrained edge devices is often overlooked in these frameworks. To tackle the computation and communication bottlenecks, \emph{federated dropout} is applied \cite{randdrop,feddrop, fjord, hermes, prunefl, adadrop}, where resource-constrained devices are allowed to train a subset of the global model, i.e. a \emph{sub-model}. Compared with a full model, the computation and communication overheads for training and transmitting sub-models are reduced, facilitating resource-constrained devices to complete the training task within constraints. 
 For example, in \cite{randdrop,feddrop}, the server randomly prunes neurons in the global model. In \cite{fjord}, the server prunes the rightmost neurons in the global model. These works fail to meet the \textit{personalization} requirement, as the server arbitrarily decides the architectures of local sub-models without considering the importance of neurons based on the \textit{local data distributions} of clients.\par
\begin{figure}
    \begin{subfigure}[b]{0.4\textwidth}
        \centering
        \includegraphics[width=\textwidth]{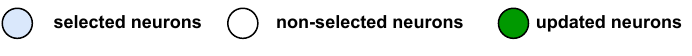}
        \label{dropoutdemo:legend}
    \end{subfigure}
    \centering
    \begin{subfigure}[b]{0.45\textwidth}
        \centering
        \includegraphics[width=\textwidth]{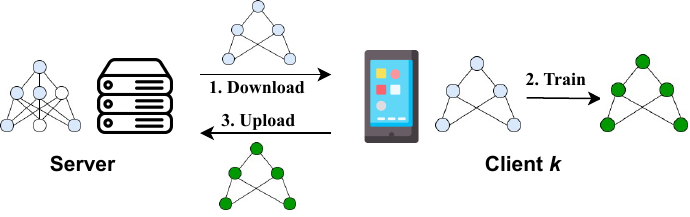}
        \caption{Dropout.}
        \label{dropoutdemo}
    \end{subfigure}
    \begin{subfigure}[b]{0.45\textwidth}
        \centering
        \includegraphics[width=\textwidth]{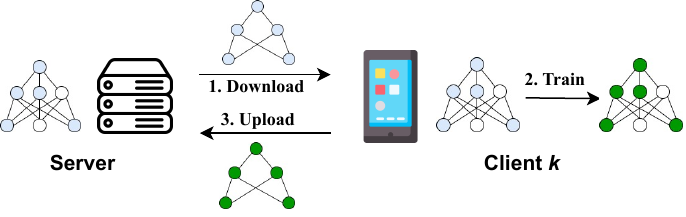}
        \caption{FedSPU.}
        \label{SPUdemo}
    \end{subfigure}
    \caption{In dropout (a), clients train sub-models with fewer parameters. In FedSPU (b), clients train full models with partial parameters frozen.}
    \label{dropoutspu}
\end{figure}
On the other hand, \cite{adadrop, hermes, prunefl} let clients adaptively prune neurons to obtain the optimal architecture of the local sub-model. Each client first pre-trains an initialized model to evaluate the importance of each neuron. Then each client locally prunes the unimportant neurons and shares the remaining sub-model with the server hereafter. Nevertheless, evaluating neuron importance requires full-model training on the client side, which might be \textit{expensive} or \textit{prohibitive} for resource-constrained devices with a critical computation bottleneck. Furthermore, the adaptive model-pruning behaviors proposed in \cite{adadrop, hermes, prunefl} rely immensely on the local data. 
The non-iid local data distributions often lead to \textit{highly unbalanced class distributions} across different clients
\cite{imbalancebias, imbalancetext, unbalanceddata, pa}. Consequently, the local dropout behavior can be \textit{heavily biased} \cite{prunefl}, and the sub-model architecture among clients may \textit{vary drastically}. In global communication, when a client absorbs parameters from other clients with inconsistent model architectures, the performance of the local model will be inevitably compromised. \par

To address the limitation of existing works, this paper introduces \textit{Federated Learning with Stochastic Parameter Update (FedSPU)}, a consolidated PFL framework aimed at mitigating the issue of \textit{local model personalization loss} while considering computation and communication bottlenecks in resource-constrained devices. It is observed that during global communication, a client's entire local sub-model is replaced by \textit{biased parameters} of other clients, leading to local model personalization loss. Therefore, if we let a client share only a \textit{partial model} with others, the adverse effect of other clients' biased parameters can be alleviated. Inspired by this, FedSPU \emph{freezes} neurons instead of pruning them, as shown in Figure \ref{dropoutspu}. Frozen neurons do not receive gradients during backpropagation and remain unaltered in subsequent updates. This approach eliminates computation overheads in backward propagation, enhancing \textit{computational efficiency}. Moreover, FedSPU does not incur extra communication overheads compared with dropout, as only the parameters of the non-frozen neurons and the positions of these neurons are communicated between clients and the server, as depicted in Figure \ref{dropoutspu}. Besides, compared with model parameters, the communication cost for sending the position indices of the non-frozen neurons is much smaller and usually ignorable \cite{hermes}. \par
Unlike pruned neurons, frozen neurons persist within a local model, and still contribute to the model's final output. This design choice incurs slightly \textit{higher computation costs} of forward propagation, but largely improves local personalization, as only a portion of a local model is replaced during communication as shown in Figure \ref{SPUdemo}. Moreover, the increased computation overhead of forward propagation is a minor problem, as in training, forward propagation constitutes a \textit{significantly smaller} portion of the total computation overhead than backpropagation \cite{vldbpytorch,he2015}. Additionally, to alleviate the overall computation and communication costs, we consolidate FedSPU with an \textit{early stopping} technique \cite{esbutwhen, flrce}. At each round, each client locally computes the training and testing errors, and compares them with the errors from the previous round. When the errors show no decrease, this client will cease training and no longer participate in FL. When all clients have halted training, FedSPU will terminate in advance to conserve computation and communication resources. 
We evaluate FedSPU on three typical deep learning datasets: EMNIST \cite{emnist}, CIFAR10 \cite{cifar} and Google Speech \cite{googlespeech}, with five state-of-the-art dropout methods included for comparison: FjORD \cite{fjord}, Hermes \cite{hermes}, FedMP \cite{adadrop}, PruneFL \cite{prunefl} and FedSelect \cite{fedselect}. Experiment results show that:
\begin{itemize}
    \item FedSPU’s unique approach of freezing neurons, rather than pruning, retains them within the local model, enhancing personalization. This design choice improves final model accuracy by an average of 4.45\% over existing dropout methods.
    \item FedSPU reduces memory usage through neuron freezing rather than full-model training, avoiding memory-intensive dropout processes. This results in significant memory savings up to 54\% at higher dropout rates.
    \item The integration of an early stopping mechanism allows FedSPU to reduce computation and communication costs significantly by 25\%\(\sim\)71\%. Compared with existing methods, FedSPU exhibits better compatibility with early stopping with an average accuracy improvement of at least 5.11\%.
\end{itemize}

\section{Related Work}\label{sec:RW}
\subsection{Personalized Federated Learning}
To the best of our knowledge, existing PFL methods can be divided into three categories. \textbf{1) Fine-tuning:} Clients at first train a global model collaboratively, followed by individual fine-tuning to adapt to local datasets. In \cite{pflayer}, each client fine-tunes some layers of the global model to adapt to the local dataset. \cite{pflmeta} proposes to find an initial global model that generalizes well to all clients through model-agnostic-meta-learning formulation, then clients fine-tune the model locally through just a few gradient steps. \cite{adaptiveFT} proposes to improve the global model's generalization with dynamic learning rate adaption based on the squared difference between local gradients, which significantly reduces the workload of local fine-tuning. \cite{clusterfl} fine-tunes the global model based on clusters, and clients from the same cluster share the same local model. \textbf{2) Personal Training:} Clients train personal models at the beginning.  \cite{pfedme} adds a regularization term to each local objective function that keeps local models from diverging too far from the global model to improve the global model's generalization. \cite{fedem} assumes each local data distribution is a mixture of several unknown distributions, and optimizes each local model with the expectation-maximization algorithm. \cite{mocha} speeds up the convergence of each local model by modeling local training as a primal-dual optimization problem. \cite{fedtha} lets each client maintain a personal head while training to improve the global model's generalization by aggregating local heads on the server side. \cite{pflod} proposes to train local models through first-order optimization, where the local objective function becomes the error subtraction between clients. \textbf{3) Hybrid:} 
% Clients are allowed to combine their local models with the global model, which helps them learn from each other and maintain personalization.
Clients merge local and global models to foster mutual learning and maintain personalization. 
In \cite{apfl, threepfl, fedala}, each client simultaneously trains the global model and its local model. The ultimate model for each client is a combination of the global and the local model. \par
Even though these works potently address the non-iid data problem in FL, they neglect the computation or communication bottleneck of resource-constrained IoT devices. 
\subsection{Dropout in Federated Learning}
In centralized machine learning, dropout is used as a regularization method to prevent a neural network from over-fitting \cite{dropout}. Nowadays, as federated learning becomes popular in the IoT industry, dropout has also been applied to address the computation and communication bottlenecks of resource-constrained IoT devices. In dropout, clients are allowed to train and transmit a subset of the global model to reduce the computation and communication overheads. To extract a subset from the global model, i.e. a sub-model, Random Dropout \cite{randdrop, feddrop} randomly prune neurons in the global model. FjORD \cite{fjord} continually prunes the right-most neurons in a neural network. FedMP \cite{adadrop}, Hermes \cite{hermes} and PruneFL \cite{prunefl} let clients adaptively prune the unimportant neurons, and the importance of a neuron is the \textit{l1-}norm, \textit{l2-}norm of parameters, and \textit{l2-}norm of gradient respectively. FedSelect \cite{fedselect} lets all clients extract a tiny sub-model first by pruning the unimportant neurons with the least gradient norms, and gradually expand the width of the sub-model during training. \par
Random Dropout and FjORD represent \textit{global dropout} methods and do not meet the personalization requirement, as the server arbitrarily prunes neurons without considering clients' non-iid data. On the other hand, \textit{local dropout} methods such as FedMP, Hermes, PruneFL and FedSelect may be hindered by the bias of unbalanced local datasets.\par 
Compared with existing works, FedSPU comprehensively meets the personalization requirement and addresses the computation and communication bottlenecks of resource-constrained devices, meanwhile overcoming data imbalance. \par

\section{FedSPU: Federated Learning with Stochastic Parameter Update}
\subsection{Problem Setting}
Given a set \(\mathbb{C} = \{1,2,...,N\} \) of clients with local datasets \(\{D_{1}, D_{2},...,D_{N}\}\) and local models \(\{w_{1},w_{2},...,w_{N}\}\). The goal of a PFL framework is to determine the optimal set of local models \(\{w^{*}_{1}, w^{*}_{2},..., w^{*}_{N}\}\) such that:
\begin{equation}\label{objeq}
    w^{*}_{1},...,w^{*}_{N} \triangleq \mathop{\arg\min}_{w_{1},...,w_{N}} \frac{1}{N} \sum_{k=1}^{N} F_{k}(w_{k})
\end{equation}
where \(w_{k}^{*}\) is the optimal model for client \(k\) (\(1\leq k \leq N\)), and \(F_{k}\) is the objective function of client \(k\). \(F_{k}\) is equivalent to the empirical risk over \(k\)'s local dataset \(D_{k}\). That is:
\begin{equation}
    F_{k}(w_{k}) = \frac{1}{n_{k}}\sum_{i=1}^{n_{k}}\mathcal{L}((\boldsymbol{x}_{i}, y_{i}), w_{k})
\end{equation}
where \(n_{k}\) is the size of dataset \(D_{k}\) and \(\mathcal{L}\) is the loss function of model \(w_{k}\) over the \(i-\)th sample (\(\boldsymbol{x}_{i}, y_{i}\)).

\begin{algorithm}
\caption{FedSPU}
\label{alg:fedspu}
\begin{algorithmic}[1]
\Require maximum global iteration \(T\), clients \(\mathbb{C}=\{1,...,N\}\), initial global model \(w^{0}\).
\State Server broadcasts \(w_{0}\) to all clients. 
\State \textbf{For} round \(t=1,2,...,T\):
\State \textbf{Server executes:}
\State \hspace{3mm} randomly sample a subset of clients \(\mathbb{C}_{t} \subset \mathbb{C}\).
\State \hspace{3mm} \(\forall k \in \mathbb{C}_{t}\):
\State \hspace{6mm} randomly sample \(A_{k}(w^{t})\) based on \(p_{k}\).
\State \hspace{6mm} send \(A_{k}(w^{t})\) to \(k\).
\State \textbf{Each client} \(k \in \mathbb{C}_{t}\) \textbf{in parallel does}:
\State \hspace{3mm} merge \(A_{k}(w^{t})\) into \(w_{k}^{t}\) to get \(\hat{w}^{t}_{k}\). \Comment{see Fig. \ref{SPUtrain}}
\State \hspace{3mm} local SGD: \(w_{k}^{t+1} = \hat{w}_{k}^{t} - \eta\nabla\bar{F}_{k}(\hat{w}_{k}^{t})\). \Comment{see Eq. (\ref{eq:SPUsgd})}
\State \hspace{3mm} send \(A_{k}(w_{k}^{t+1})\) to the server. \Comment{see Fig. \ref{SPUtrain}}
\State \textbf{Server executes:}
\State \hspace{3mm} \(\forall k \in \mathbb{C}_{t}\):
\State \hspace{6mm} receive \(A_{k}(w_{k}^{t+1})\) from \(k\).
\State \hspace{3mm} Aggregate all \(A_{k}(w_{k}^{t+1})\) to get \(w^{t+1}\). \Comment{see Fig. \ref{SPUaggregate}}
\State \Return \(w_{1},w_{2},...,w_{N}\).
\end{algorithmic}
\end{algorithm}

\subsection{Solution Overview}
\begin{figure}
    \centering
    \begin{subfigure}[b]{0.45\textwidth}
        \centering
        \includegraphics[width=\textwidth]{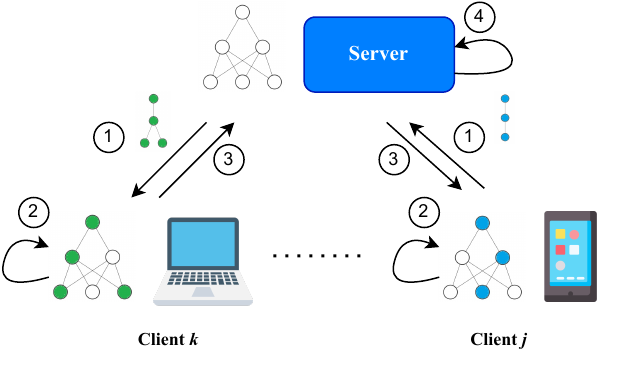}
        \caption{Overview of FedSPU.}
        \label{SPUoverview}
    \end{subfigure}
    \begin{subfigure}[b]{0.45\textwidth}
        \centering
        \includegraphics[width=\textwidth]{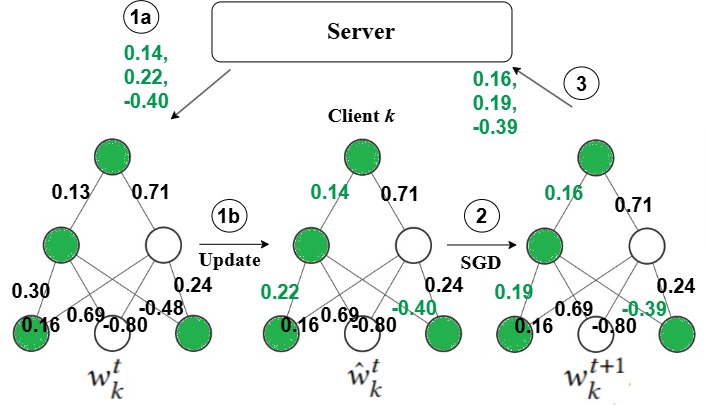}
        \caption{The local training procedure of client \(k\) in FedSPU.}
        \label{SPUtrain}
    \end{subfigure}
    \caption{Demonstration of the FedSPU framework.}
    \label{spudesign}
\end{figure}

A comprehensive procedure of FedSPU is presented in Algorithm \ref{alg:fedspu}, with an overview presented in Figure \ref{spudesign}. As shown in Figure \ref{SPUoverview}, at round \(t\), the server randomly selects a set of participating clients \(\mathbb{C}_{t}\) and executes steps \textcircled{1}\(\sim\)\textcircled{4}. \par

\textcircled{1}. For every participating client \(k\), the server selects a set of active neurons from the global model, and sends the neurons' parameters \(A_{k}(w^{t})\) to \(k\). Specifically, in each layer, random \(p_{k}\) of the neurons are selected, where \(p_{k}\in (0,1]\) is the ratio of active neurons. The value of \(p_{k}\) depends on the system characteristic of the client \(k\), with more powerful \(k\)'s device (e.g. base station, data silo) having larger \(p_{k}\). The active neurons are selected randomly to ensure uniform parameter updates.
Locally, client \(k\) updates the local model \(w_{k}^{t}\) with the received \(A_{k}(w^{t})\) to obtain an intermediate model \(\hat{w}_{k}^{t}\) as shown in Figure \ref{SPUtrain}. 
\par
\textcircled{2}. Client \(k\) updates model \(\hat{w}_{k}^{t}\) using stochastic gradient descent (SGD) to get a new model \(w_{k}^{t+1}\) following Equation (\ref{eq:SPUsgd}):
\begin{equation}\label{eq:SPUsgd}
    w_{k}^{t+1} = \hat{w}_{k}^{t} - \eta\nabla\bar{F}_{k}(\hat{w}_{k}^{t})
\end{equation}
where \(\eta\) is the learning rate and \(\nabla\bar{F}_{k}(\hat{w}_{k}^{t})\) is the gradient of \(F_{k}\) with respect to only the active parameters. That is, for all elements \{\(\hat{w}_{k,1}^{t},...,\hat{w}_{k,m}^{t}\)\} in \(\hat{w}_{k}^{t}\), we have:
\begin{equation}\label{eq:barfk}
         \nabla \bar{F}_{k}(\hat{w}^{t}_{k,i}) = 
         \begin{cases}
         \nabla F_{k}(\hat{w}^{t}_{k,i}), & \hat{w}^{t}_{k,i} \in A_{k}(w^{t}) \\
         0, & \text{otherwise.}
         \end{cases},
1 \leq i \leq m
\end{equation}
In this step, only the active parameters are updated as shown in Figure \ref{SPUtrain}.\par 
\textcircled{3}. Client \(k\) uploads the updated active parameters \(A_{k}(w_{k}^{t+1})\) to the server. \par

\begin{figure}[ht]
    \centering
    \includegraphics[scale=0.45]{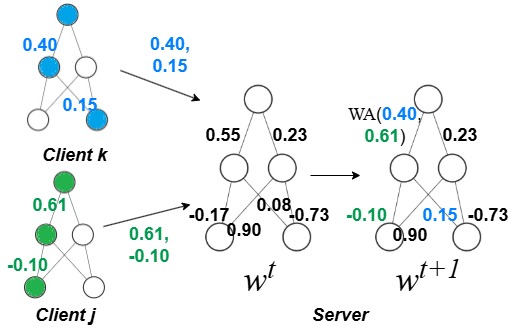}
    \caption{The aggregation scheme in FedSPU.}
    \label{SPUaggregate}
\end{figure}

\textcircled{4}. The server aggregates all updated parameters and updates the global model. In this step, FedSPU applies a standard aggregation scheme commonly used in existing dropout methods, where only the active parameters get aggregated and updated \cite{randdrop, fjord, hermes}. Figure \ref{SPUaggregate} shows a simple example of how the aggregation scheme works, where "WA" stands for weighted average. \par

\subsection{Full Local Model Preserves Personalization}
\begin{figure}
    \centering
    \includegraphics[scale=0.6]{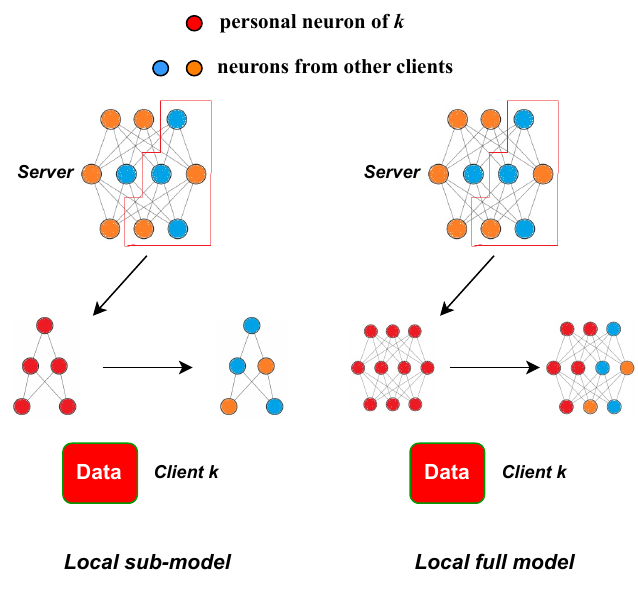}
    \caption{Comparison between replacing all parameters of a local sub-model and replacing a portion of the parameters for a local full model during global communication.}
    \label{moti2}
\end{figure}

FedSPU freezes neurons instead of pruning them to preserve the integrity of the local model architecture, thereby preserving the personalization of local models. For clarity, Figure \ref{moti2} shows a comparison between a local sub-model and a local full model. As shown in the left-hand side of Figure \ref{moti2}, in global communication, when receiving other clients' biased parameters from the server, the entire local sub-model is replaced, resulting in a loss of personalization. Conversely, as illustrated in the right-hand side of Figure \ref{moti2}, for a local full model, only partial parameters are replaced, while the remainder remains personalized. This limits the adverse effect of biased parameters from other clients, enabling the local model to maintain performance on the local dataset. \par

\subsection{Enhancing FedSPU with Early Stopping Strategy}
Since FedSPU slightly increases the computation overhead, it requires more
computation resources (e.g., energy \cite{flRCsurvey}, time \cite{he2015})
for training. This may pose a challenge for resource-constrained devices. 
To address this concern, it is expected to reduce the training time of FedSPU without sacrificing accuracy \cite{flrce}. Motivated by this, we enhance FedSPU with the \textbf{Early Stopping (ES)} technique \cite{esbutwhen} to prevent clients from unnecessary training to avoid the substantial consumption of computation and communication resources. At round \(t\), after training, each client \(k\) computes \(\mathcal{L}_{t}\) following Equation (\ref{eq:localloss}):
\begin{equation}\label{eq:localloss}
    \mathcal{L}_{t} = \lambda\mathcal{L}_{train} + (1-\lambda)\mathcal{L}_{test}.
\end{equation}
\(\mathcal{L}_{train}\) is the training error of current round \(t\), \(\mathcal{L}_{test}\) is the testing error of \(w_{k}^{t}\) on \(k\)'s validation set. \(\lambda \in (0,1)\) is the train-test split factor of the local dataset \(D_{k}\). When the loss \(\mathcal{L}_{t}\) is non-decreasing, i.e. \(\mathcal{L}_{t}>\mathcal{L}_{t-1}\), client \(k\) will stop training and no longer participate in FL due to
resource concerns. If all clients have stopped training before the maximum global iteration \(T\), FedSPU will terminate prematurely. The enhanced FedSPU framework with early stopping is detailed in Algorithm \ref{alg:fedspuES}.
\begin{algorithm}
\caption{FedSPU with Early Stopping (FedSPU + ES)}
\label{alg:fedspuES}
\begin{algorithmic}[1]
\Require maximum global iteration \(T\), clients \(\mathbb{C}=\{1,...,N\}\), initial global model \(w^{0}\).
\State Server broadcasts \(w_{0}\) to all clients. 
\State \textbf{For} round \(t=1,2,...,T\):
\State \textbf{Server executes:}
\State \hspace{3mm} randomly sample a subset of clients \(\mathbb{C}_{t} \subset \mathbb{C}\).
\State \hspace{3mm} \(\forall k \in \mathbb{C}_{t}\):
\State \hspace{6mm} randomly sample \(A_{k}(w^{t})\) based on \(p_{k}\).
\State \hspace{6mm} send \(A_{k}(w^{t})\) to \(k\).
\State \textbf{Each client} \(k \in \mathbb{C}_{t}\) \textbf{in parallel does}:
\State \hspace{3mm} merge \(A_{k}(w^{t})\) into \(w_{k}^{t}\) to get \(\hat{w}^{t}_{k}\). \Comment{see Fig. \ref{SPUtrain}}
\State \hspace{3mm} local SGD: \(w_{k}^{t+1} = \hat{w}_{k}^{t} - \eta\nabla\bar{F}_{k}(\hat{w}_{k}^{t})\). \Comment{see Eq. (\ref{eq:SPUsgd})}
\State \hspace{3mm} compute \(\mathcal{L}_{t}\). \Comment{see Eq. (\ref{eq:localloss})}
\State \hspace{3mm} \textbf{If} \(\mathcal{L}_{t} > \mathcal{L}_{t-1}\):
\State \hspace{6mm} \textit{status} \(\gets\) \textit{stopped}
\State \hspace{3mm} \textbf{Else:}
\State \hspace{6mm} \textit{status} \(\gets\) \textit{on}
\State \hspace{3mm} send \(A_{k}(w_{k}^{t+1})\) and \textit{status} to the server. 
\State \textbf{Server executes:}
\State \hspace{3mm} \(\forall k \in \mathbb{C}_{t}\):
\State \hspace{6mm} receive \(A_{k}(w_{k}^{t+1})\), \textit{status} from \(k\).
\State \hspace{6mm} \textbf{If} \textit{status} == \textit{stopped}:
\State \hspace{9mm} remove \(k\) from \(\mathbb{C}\).
\State \hspace{3mm} Aggregate all \(A_{k}(w_{k}^{t+1})\) to get \(w^{t+1}\). \Comment{see Fig. \ref{SPUaggregate}}
\State \hspace{3mm} \textbf{If} \(\mathbb{C}==\varnothing\):
\State \hspace{6mm} \textbf{TERMINATE}.
\State \Return \(w_{1},w_{2},...,w_{N}\).
\end{algorithmic}
\end{algorithm}

\begin{table*}
    \centering
    \begin{tabular}{c|c|c|c|c|c|c}
    \hline
        \textbf{Dataset} & PruneFL & FjORD & Hermes & FedMP & FedSelect & FedSPU \\
    \hline
        EMNIST  & \(67.34\pm0.9\) & \(7.66\pm0.4\) & \(69.09\pm0.6\) & \(67.42\pm0.9\) & \(66.26\pm1.7\) & \textbf{73.42\(\pm0.4\)} \\
        CIFAR10  & \(36.65\pm2.0\) & \(24.95\pm2.2\) & \(40.52\pm4.7\) & \(33.48\pm1.5\) & \(47.83\pm1.1\) & \textbf{51.81}\(\pm1.5\) \\
        Google Speech  & \(21.5\pm2.0\) & \(11.20\pm7.0\) & \(32.03\pm3.0\) & \(21.08\pm1.7\) & \(34.06\pm2.1\) & \textbf{39.1}\(\pm2.8\) \\
    \hline
    \end{tabular}
    \caption{Mean final test accuracy (\%) across three Dirichlet distributions with parameters 0.1, 0.5 and 1.0 (\textbf{without ES}).}
    \label{tab:final acc}
\end{table*}

\begin{table*}
    \centering
    \begin{tabular}{c|c|c|c|c|c|c}
    \hline
        \textbf{Dataset} & PruneFL & FjORD & Hermes & FedMP & FedSelect & FedSPU \\
    \hline
        EMNIST  & \(62.8\pm3.8\) & \(0.08\pm0.3\) & \(68.83\pm1.2\) & \(63.0\pm2.7\) & \(62.3\pm4.9\) & \textbf{73.31\(\pm0.2\)} \\
        CIFAR10  & \(31.2\pm2.1\) & \(21.2\pm3.2\) & \(30.6\pm3.3\) & \(26.9\pm0.5\) & \(37.85\pm0.3\) & \textbf{42.66}\(\pm1.5\) \\
        Google Speech  & \(16.7\pm2.0\) & \(10.68\pm10.2\) & \(29.66\pm2.0\) & \(16.6\pm2.2\) & \(19.2\pm2.6\) & \textbf{35.7}\(\pm1.6\) \\
    \hline
    \end{tabular}
    \caption{Mean test accuracy (\%) across three Dirichlet distributions with parameters 0.1, 0.5 and 1.0 (\textbf{with ES}).}
    \label{tab: ACC ES}
\end{table*}

\section{Convergence Analysis}\label{sec:theo}
This section analyzes the convergence of each local model in FedSPU. We first make some common assumptions following existing works \cite{hermes, adadrop}: \par
\textbf{Assumption 1.} \textit{Every local objective function \(F_{k}\) is \(L-\)smooth (\(L>0\)). That is, \(\forall w_{1}, w_{2}\), we have:} 

\[F_{k}(w_{2})-F_{k}(w_{1}) \leq \langle\nabla F_{k}(w_{1}), w_{2}-w_{1}\rangle + \frac{L}{2}\|w_{2}-w_{1}\|^{2}.\]
\par
\textbf{Assumption 2.}
\textit{The divergence between local gradients with and without incorporating the parameter received from the server is bounded:}\[\exists \ Q > 0, \forall \ k,t, \ \frac{\mathbb{E}(\|\nabla F_{k}(w_{k}^{t})\|^{2})}{\mathbb{E}(\|\nabla F_{k}(\hat{w}_{k}^{t})\|^{2})} \leq Q.\]\par
\textbf{Assumption 3.}
\textit{The divergence between the local parameters with and without incorporating the parameter received from the server is bounded:} \[\exists \ \sigma > 0, \forall \ k, t,\ \mathbb{E}(\|\hat{w}_{k}^{t} - w_{k}^{t}\|^2) \leq \sigma^{2}.\]\par 
 Additionally, with respect to the gradient \(\nabla\bar{F}_{k}\), we derive the following lemmas: \par
\textbf{Lemma 1.} \(\forall \ w_{k}, \ \mathbb{E}(\|\bar{F}_{k}(w_{k})\|^{2}) = p_{k}^{2} \ \|F_{k}(w_{k})\|^{2}\). \par
\textbf{Lemma 2}. \(\forall w_{k}, \langle \nabla F_{k}(w_{k}), \nabla \bar{F}_{k}(w_{k}) \rangle = \|\nabla \bar{F}_{k}(w_{k})\|^{2}\).\par 
Based on the assumptions and lemmas, Theorem 1 holds:\par
\textbf{Theorem 1.} \textit{When the learning rate \(\eta\) satisfies \(\eta < \frac{1+\sqrt{1-\frac{QL}{p_{k}^{2}}}}{L}\), every local model \(w_{k}\) will at least reach a \(\epsilon\)\emph{-critical point} \(w_{k}^{\epsilon}\) (i.e. \(\|\nabla F_{k}(w_{k}^{\epsilon})\| \leq \epsilon\)) in \(O(\frac{w_{k}^{0}-w_{k}^{\epsilon}}{\epsilon\eta})\) rounds, with \(\epsilon = \sqrt{\frac{(L+1)Q\sigma^{2}}{(2\eta-L\eta^{2})p_{k}^{2}+Q}}\).} \par
According to Theorem 1, in FedSPU, each client's local objective function will 
% at least decrease to a relatively small value, 
converge to a relatively low value,
given that the learning rate is small enough. This means that every client's personal model will eventually acquire favorable performance on the local dataset even if the objective function is not necessarily convex. The proofs can be found in the Appendix. \par

\begin{table*}[ht]
    \centering
    \begin{tabular}{c|c|c|c|c|c|c|c}
    \hline
    \multicolumn{8}{c}{\textbf{Total time of local training (in hours)}} \\
    \hline
        \textbf{Dataset} & PruneFL & FjORD & Hermes & FedMP & FedSelect & FedSPU & FedSPU+ES \\
    \hline
        EMNIST & 8.11 & 7.85 & 8.57 & 7.30 & 7.68 & 7.82 & \textbf{4.0} \\
        CIFAR10  & 24.69 & 25.03 & 25.18 & 25.5 & 25.06 & 25.05 & \textbf{8.9} \\
        Google Speech & 8.29 & 7.83 & 8.09 & 8.16 & 8.03 & 8.71 & \textbf{5.97} \\
    \hline
    \multicolumn{8}{c}{\textbf{Total size of parameter transmission (in GB)}} \\
    \hline
        EMNIST & 11.69 & 11.72 & 11.73 & 11.66 & 20.6 & 11.71 & \textbf{7.23} \\
        CIFAR10  & 18.20 & 18.17 & 18.2 & 18.43 & 20.7 & 18.04 & \textbf{6.37} \\
        Google Speech & 4.39 & 4.36 & 4.36 & 4.36 & 8.39 & 4.36 & \textbf{2.99} \\
    \hline
    \end{tabular}
    \caption{Comparison of the total training time (hours) and the size of transmitted parameters for \(T=500\) rounds. }
    \label{tab:cost}
\end{table*}

\section{Experiment}\label{sec:exp}
\subsection{Experiment Setup}\label{sec:setup}
\textbf{Datasets and models.} We evaluate FedSPU on three real-world datasets that are very commonly used in the state-of-the-art, including: \textbf{Extended MNIST (EMNIST)} contains 814,255 images of human-written digits/characters from 62 categories (numbers 0-9 and 52 upper/lower-case English letters). Each sample is a black-and-white-based image with \(28\times28\) pixels \cite{emnist}. \textbf{CIFAR10} contains 50,000 images of real-world objects across 10 categories. Each sample is an RGB-based colorful image with \(32\times 32\) pixels \cite{cifar}. \textbf{Google Speech} is an audio dataset containing 101,012 audio commands from more than 2,000 speakers. Each sample is a human-spoken word belonging to one of the 35 categories \cite{googlespeech}.
For EMNIST and Google Speech, a convolutional neural network (CNN) with two convolutional layers and one fully-connected layer is used, following the setting of \cite{fjord}. For CIFAR10, a CNN with two convolutional layers and three fully-connected layers is used, following the setting of \cite{hermes}.\par
We conduct three runs on each dataset. For each run, data are allocated to clients unevenly following the settings of \cite{feddyn, nofear}, following a \textit{Dirichlet distribution} with parameter \(\alpha\). We tune the value of \(\alpha\) with 0.1, 0.5 and 1.0 to create three different distributions for each run. We split each client's dataset into a training set and a testing set with the split factor \(\lambda=0.7\). \par

\textbf{Baselines.}
We compare FedSPU with five typical methods: \textbf{FjORD \cite{fjord}:} The server prunes neurons in a fixed right-to-left order. \textbf{FedSelect \cite{fedselect}:} All clients start with a small sub-model and gradually expand it. \textbf{FedMP \cite{adadrop}:} Each client \(k\) locally prunes neurons to create a personal sub-model. In each layer, \(1-p_{k}\) of the neurons with the least importance scores are pruned. The importance of a neuron is defined as the \textit{l1-}norm of the parameters. \textbf{Hermes \cite{hermes}:} Similar to FedMP, each client \(k\) locally prunes the \(1-p_{k}\) least important neurons in each layer. The importance of a neuron is defined as the \textit{l2-}norm of the parameters. \textbf{PruneFL \cite{prunefl}:} Similarly, each client \(k\) locally prunes the \(1-p_{k}\) least important neurons in each layer. The importance of a neuron is defined as the \textit{l2-}norm of the neuron's gradient. \par

\begin{table}
    \centering
    \begin{tabular}{c|cc}
    \hline
         \textbf{Dataset} & \multicolumn{2}{|c}{\textbf{Rounds}} \\
         &FedSPU&FedSPU+ES\\
    \hline
         EMNIST (\(\alpha=0.1\)) &500&207 \\
         EMNIST (\(\alpha=0.5\)) &500&374 \\
         EMNIST (\(\alpha=1.0\)) &500&346 \\
    \hline
         CIFAR10 (\(\alpha=0.1\)) &500&171 \\
         CIFAR10 (\(\alpha=0.5\)) &500&216 \\
         CIFAR10 (\(\alpha=1.0\)) &500&147 \\
    \hline
        Google Speech (\(\alpha=0.1\)) &500&306 \\
        Google Speech (\(\alpha=0.5\)) &500&358 \\
        Google Speech (\(\alpha=1.0\)) &500&365 \\
    \hline
    \end{tabular}
    \caption{Comparison of final accuracy and estimated computation and communication cost (combined) between FedSPU and FedSPU with early stopping. 
    }
    \label{tab:es comprae}
\end{table}

\textbf{Parameter settings and system implementation.} The maximum global iteration is set to \(T=500\) with a total of \(M=100\) clients. The number of active clients per round is set to 10, and each client has five local training epochs \cite{fjord}. For ES, if the number of non-stopped clients is less than 10, then all non-stopped clients will be selected. The learning rate is set to 2e-4, 5e-4 and 0.1 respectively for EMNIST, Google Speech and CIFAR10. The batch size is set to 16 for EMNIST and Google Speech, and 128 for CIFAR10. The experiment is implemented with Pytorch 2.0.0 and the Flower framework \cite{flwr}. The server runs on a desktop computer and clients run on NVIDIA Jetson Nano Developer Kits with one 128-core Maxwell GPU and 4GB 64-bit memory. For the emulation of system heterogeneity and resource constraints, we divide the clients into 5 uniform clusters following \cite{fjord}. Clients of the same cluster share the same value of \(p_{k}\). The values of \(p_{k}\) for the five clusters are 0.2, 0.4, 0.6, 0.8 and 1.0 respectively. For FedSelect, the initial and final values of \(p_{k}\) are set to 0.25 and 0.5 \cite{fedselect}. \par

\subsection{Experiment Results}
\textbf{Accuracy.} FedSPU obtains higher final accuracy than dropout as Table \ref{tab:final acc} shows. On average, FedSPU improves the final test accuracy by \(4.45\%\) compared with the best results of dropout (Hermes). These results prove the usefulness of local full models in preserving personalization. \par

\textbf{Computation and communication overheads.} To assess computation overhead, we focus on the wall-clock training time rather than floating point operations out of practical concern \cite{feddesign,he2015,pyramid}. As shown by Table \ref{tab:cost}, the additional computation overhead caused by FedSPU is minor. Among all cases, the training time of FedSPU is less than \(1.11\times\) that of the fastest baseline. Moreover, FedSPU does not incur extra communication overhead compared with dropout. As shown in Table \ref{tab:cost}, there is very little difference between the size of the transmitted parameters in FedSPU and dropout. \par 

\begin{figure*}
    \begin{subfigure}[b]{0.312\textwidth}
        \centering
        \includegraphics[width=\textwidth,
        ]{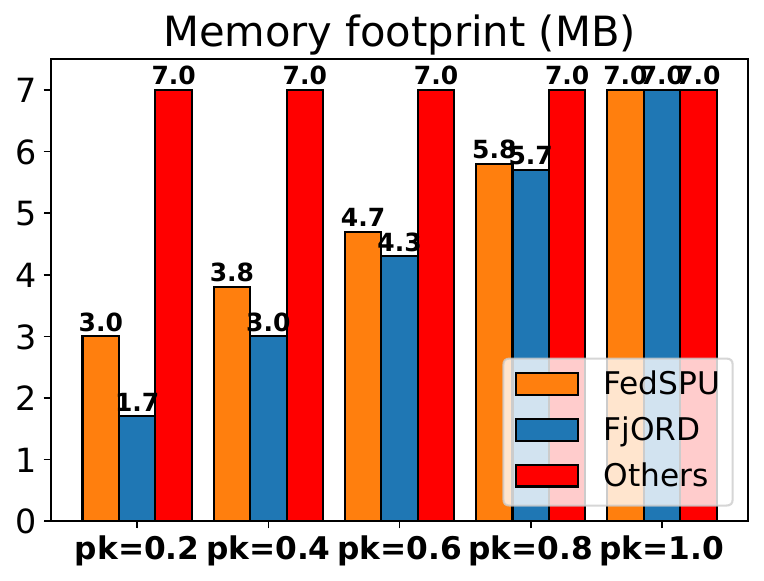}
        \caption{EMNIST.}
        \label{emnist energy}
    \end{subfigure}
    \begin{subfigure}[b]{0.32\textwidth}
        \centering
        \includegraphics[width=\textwidth,
        ]{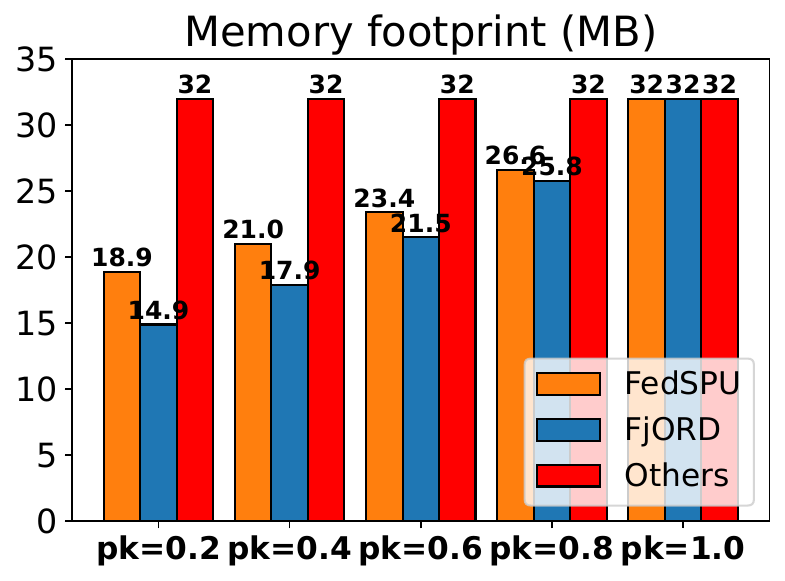}
        \caption{CIFAR10.}
        \label{cifar energy}
    \end{subfigure}
    \begin{subfigure}[b]{0.32\textwidth}
        \centering
        \includegraphics[width=\textwidth,
        ]{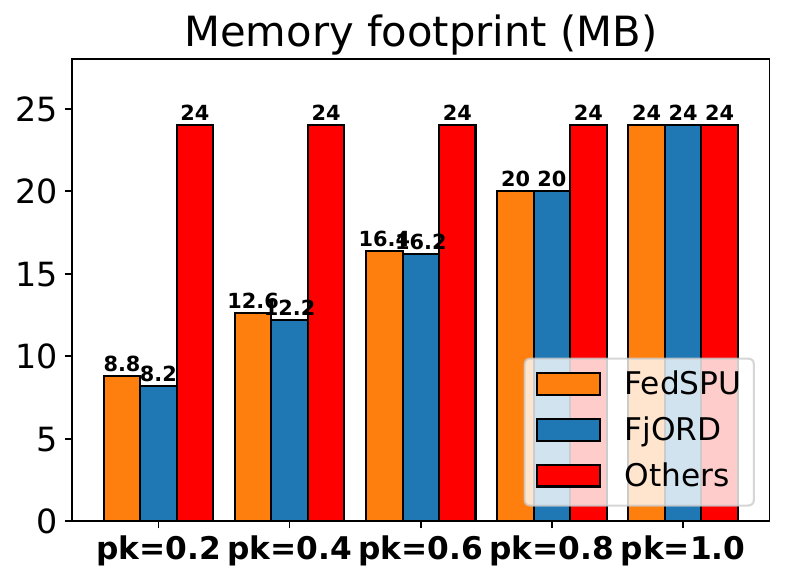}
        \caption{Google Speech.}
        \label{voice energy}
    \end{subfigure}
    \caption{Comparison of memory footprint (MB) with different \(p_{k}\). "Others" stand for FedMP, Hermes, PruneFL and FedSelect.}
    \label{img:memory}
\end{figure*}

\textbf{Effect of the early stopping strategy.} 
With early stopping (ES), the number of training rounds of FedSPU is reduced by \(25\%\sim71\%\) as shown in Table \ref{tab:es comprae}. Correspondingly, the total computation and communication overheads are also reduced as shown in Table \ref{tab:cost}. \par 
As shown in Tables \ref{tab:final acc} and \ref{tab:es comprae}, for EMNIST and Google Speech, ES effectively alleviates the computation/communication cost by \(25\%\sim59\%\) in FedSPU with a marginal accuracy sacrifice of 0.11\% and 3.4\%. For CIFAR10, the ES strategy becomes more aggressive, reducing the cost by \(57\%\sim71\%\) with \(9.2\%\) of accuracy loss. Despite this, the final accuracy of FedSPU+ES is still higher than that of dropout in most cases as shown in Tables \ref{tab:final acc} and \ref{tab: ACC ES}. \par

For fairness, we also evaluate the baselines' performance with ES. As shown in Table \ref{tab: ACC ES}, FedSPU consistently obtains the highest accuracy, demonstrating better compatibility with the ES mechanism. Overall, FedSPU improves the mean accuracy by at least 5.11\% in the presence of ES.

\textbf{Memory footprint.} We calculate the memory footprint by \cite{slt}, where the total memory footprint equals the accumulated size of weights, gradients, and activations stored in the model. As Figure \ref{img:memory} shows, FedSPU significantly reduces the memory footprint compared with FedSelect, PruneFL, FedMP and Hermes ("Others" in Figure \ref{img:memory}) which require full-model training, achieving average 54\%, 44\%, 31\% and 18\% reduction with \(p_{k}=0.2,0.4,0.6,0.8\) respectively.  \par

\section{Conclusion}
We propose FedSPU, a novel personalized federated learning approach with stochastic parameter update. FedSPU preserves the global model architecture on each edge device, randomly freezing portions of the local model based on device capacity, training the remaining segments with local data, and subsequently updating the model based solely on the trained segments. This methodology ensures that a segment of the local model remains personalized, thereby mitigating the adverse effects of biased parameters from other clients. We also propose to combine FedSPU with early stopping to mitigate the training iterations, which further reduces the overall computation and communication costs while maintaining high accuracy. In the future, we plan to explore the similarities of local clients in a privacy-preserving way, leveraging techniques such as learning vector quantization \cite{qin2005initialization} and graph matching \cite{gong2016discrete} to guide the model freezing process and enhance local model training. Furthermore, we intend to extend FedSPU to traditional FL problems, and enhance the generalization capability of the global model.

\section{Acknowledgements}

This work is funded by the Australian Research Council under Grant No. DP220101823, DP200102611, and LP180100114.

\bibliography{aaai25}

\appendix
\section*{Theoretical Proof}
This section shows the detailed proof of the paper's convergence analysis, first, we re-write all assumptions and lemmas: \par
\hfill
\par
\textbf{Assumption 1.} \textit{Every local objective function \(F_{k}\) is \(L-\)smooth:} \[\forall w_{1}, w_{2}, F_{k}(w_{2})-F_{k}(w_{1}) \leq \langle\nabla F_{k}(w_{1}), w_{2}-w_{1}\rangle + \frac{L}{2}\|w_{2}-w_{1}\|^{2}.\]\par
\textbf{Assumption 2.}
\textit{The divergence between local gradients with and without incorporating the parameter received from the server is bounded:}\[\exists \ Q > 0, \forall \ k,t, \ \frac{\mathbb{E}(\|\nabla F_{k}(w_{k}^{t})\|^{2})}{\mathbb{E}(\|\nabla F_{k}(\hat{w}_{k}^{t})\|^{2})} \leq Q.\]\par
\textbf{Assumption 3.}
\textit{The divergence between the local parameters with and without incorporating the parameter received from the server is bounded:} \[\exists \ \sigma > 0, \forall \ k, t,\ \mathbb{E}(\|\hat{w}_{k}^{t} - w_{k}^{t}\|^2) \leq \sigma^{2}.\]\par
\hfill
\par
 With respect to the gradient \(\nabla\bar{F}_{k}\), we derive the following lemmas: \par
 \hfill
 \par
\textbf{Lemma 1.} \(\forall \ w_{k}, \ \mathbb{E}(\|\bar{F}_{k}(w_{k})\|^{2}) = p_{k}^{2} \ \|F_{k}(w_{k})\|^{2}\). \par
\hfill
\par
\textit{Proof.} As defined in FedSPU, a parameter \(w_{k,i}\) in \(w_{k}\) is active only when the two neurons it connects are both active, and the probability of the two neurons being both active is \(p_{k}^{2}\). Therefore:
\begin{equation}\label{eq:l1proof}
     \begin{split}
         & \mathbb{E}(\|\nabla \bar{F}_{k}(w_{k})\|^2) \\
         & = \mathbb{E}(\sum_{i}^{n} \nabla \bar{F}_{k}(w_{k,i})^{2}) \\
         & = \sum_{i}^{n} \mathbb{E}(\nabla \bar{F}_{k}(w_{k,i})^{2}) \\
         & = p_{k}^{2}(\nabla F_{k}(w_{k,1})^{2} + \nabla F_{k}(w_{k,2})^{2} + ... + \nabla F_{k}(w_{k,m})^{2}) \\
         & = p_{k}^{2} \ \|\nabla F_{k}(w_{k})\|^{2} .\\
     \end{split}  
\end{equation}\par
\textbf{Lemma 2}. \(\forall w_{k}, \langle \nabla F_{k}(w_{k}), \nabla \bar{F}_{k}(w_{k}) \rangle = \|\nabla \bar{F}_{k}(w_{k})\|^{2}\).\par 
\textit{Proof.}
\begin{equation}
    \begin{split}
        & \hspace{5mm} \langle \nabla F_{k}(w_{k}), \nabla \bar{F}_{k}(w_{k}) \rangle \\
        & = \sum_{i} F_{k}(w_{k,i}) \ \bar{F}_{k}(w_{k,i}) \\
        & = \sum_{i, i \not\in A(w_{k})}F_{k}(w_{k,i}) \ \bar{F}_{k}(w_{k,i}) + \sum_{i, i \in A(w_{k})}F_{k}(w_{k,i}) \ \bar{F}_{k}(w_{k,i}) \\
        & =  0 + \sum_{i, i \in A(w_{k})}F_{k}(w_{k,i}) \ \bar{F}_{k}(w_{k,i}) \hspace{15mm} \\
        & = \sum_{i, i \in A(w_{k})}F_{k}(w_{k,i})^{2} \\
        & = \|\nabla \bar{F}_{k}(w_{k})\|^{2} .\\
    \end{split}
\end{equation}
Based on Assumptions 1-3 and Lemmas 1 and 2, we derive Theorem 1:\par
\textbf{Theorem 1.} \textit{When the learning rate \(\eta\) satisfies \(\eta < \frac{1+\sqrt{1-\frac{QL}{p_{k}^{2}}}}{L}\), every local model \(w_{k}\) will at least reach a \(\epsilon\)\emph{-critical point} \(w_{k}^{\epsilon}\) (i.e. \(\|\nabla F_{k}(w_{k}^{\epsilon})\| \leq \epsilon\)) in \(O(\frac{w_{k}^{0}-w_{k}^{\epsilon}}{\epsilon\eta})\) rounds, with \(\epsilon = \sqrt{\frac{(L+1)Q\sigma^{2}}{(2\eta-L\eta^{2})p_{k}^{2}+Q}}\).} \par

\textit{Proof.} As \(F_{k}\) is \(L-smooth\), we have:
\begin{equation}\label{smooth1}
    \begin{split}
       & F_{k}({w}_{k}^{t+1}) - F_{k}(\hat{w}_{k}^{t}) \\ 
       & \leq \langle \nabla F_{k}(\hat{w}_{k}), w_{k}^{t+1} - \hat{w}_{k}^{t} \rangle + \frac{L}{2}\|w_{k}^{t+1} - \hat{w}_{k}^{t}\|^{2} \\ 
       & = - \eta \|\nabla \bar{F}_{k}(\hat{w}_{k}^{t})\|^{2} + \frac{L\eta^{2}}{2} \|\nabla \bar{F}_{k}(\hat{w}_{k}^{t})\|^{2} \\
       & = (-\eta + \frac{L\eta^{2}}{2}) \|\nabla \bar{F}_{k}(\hat{w}_{k}^{t})\|^{2}. \\
    \end{split}
\end{equation}
and:
\begin{equation}\label{smooth2}
    \begin{split}
       & F_{k}(\hat{w}_{k}^{t}) - F_{k}({w}_{k}^{t}) \\ & \leq \langle \nabla F_{k}(w_{k}), \hat{w}_{k}^{t} - w_{k}^{t} \rangle + \frac{L}{2}\|\hat{w}_{k}^{t} - w_{k}^{t}\|^{2} \\
       & \leq \frac{1}{2} \|\nabla F_{k}(w_{k}^{t})\|^{2} + \frac{1}{2} \|\hat{w}_{k}^{t}-w_{k}^{t}\|^{2} + \frac{L}{2}\|\hat{w}_{k}^{t} - w_{k}^{t}\|^{2}. \\
    \end{split}
\end{equation}
\par
\hfill
\par
By adding (\ref{smooth1}) and (\ref{smooth2}), we obtain:
\begin{equation}
    \begin{split}
        & F_{k}(w_{k}^{t+1}) - F_{k}({w}_{k}^{t}) \\
        & \leq (-\eta + \frac{L\eta^{2}}{2}) \|\nabla \bar{F}_{k}(\hat{w}_{k}^{t})\|^{2} + \frac{1}{2}\|\nabla F_{k}(w_{k}^{t})\|^{2} + \frac{L+1}{2} \|\hat{w}_{k}^{t}-w_{k}^{t}\|^{2}. \\
    \end{split}
\end{equation}
\par By taking the expectation on both sides, we obtain:
\begin{equation}
    \begin{split}
        & \mathbb{E}(F_{k}(w_{k}^{t+1}) - F_{k}({w}_{k}^{t})) \\
        & \leq (-\eta + \frac{L\eta^{2}}{2}) \ \mathbb{E}( \|\nabla \bar{F}_{k}(\hat{w}_{k}^{t})\|^{2}) \\
        & \hspace{5mm} + \frac{1}{2}\mathbb{E}(\|\nabla F_{k}(w_{k}^{t})\|^{2}) + \frac{L+1}{2} \mathbb{E}(\|\hat{w}_{k}^{t}-w_{k}^{t}\|^{2}). \\
        & \leq (-\eta + \frac{L\eta^{2}}{2}) \ \mathbb{E}( \|\nabla \bar{F}_{k}(\hat{w}_{k}^{t})\|^{2}) \\
        & \hspace{5mm} \ + \frac{1}{2}\mathbb{E}(\|\nabla F_{k}(w_{k}^{t})\|^{2}) + \frac{L+1}{2} \sigma^{2}. \\
    \end{split}
\end{equation}
Since \(\frac{1}{2} \mathbb{E}(\|\nabla F_{k}(w_{k}^{t})\|^{2}) + \frac{L+1}{2}\sigma^{2} \) is always positive, in order for \(F_{k}\) to decrease, we need \((-\eta + \frac{L\eta^{2}}{2})\ \mathbb{E}( \|\nabla \bar{F}_{k}(\hat{w}_{k}^{t})\|^{2})\) to be less than 0, i.e. \(\eta < \frac{2}{L}\).\par
\hfill
\par
Furthermore, based on Lemma 1 and Assumption 2, we have \(\mathbb{E}( \|\nabla \bar{F}_{k}(\hat{w}_{k}^{t})\|^{2}) = p_{k}^{2} \ \mathbb{E}( \|\nabla F_{k}(\hat{w}_{k}^{t})\|^{2}) \geq  \frac{1}{Q} \ p_{k}^{2} \ \mathbb{E}( \|\nabla F_{k}(w_{k}^{t})\|^{2}) \). Therefore:
\begin{equation}\label{smooth3}
    \begin{split}
        & \mathbb{E}(F_{k}(w_{k}^{t+1}) - F_{k}({w}_{k}^{t})) \\
        & \leq ((-\eta + \frac{L\eta^{2}}{2})\frac{p_{k}^{2}}{Q} + \frac{1}{2}) \ \mathbb{E}( \|\nabla F_{k}(w_{k}^{t})\|^{2}) + \frac{L+1}{2}\sigma^{2}. \\
    \end{split}
\end{equation}
\par
\hfill
\par
When \((-\eta + \frac{L\eta^{2}}{2}) \frac{p_{k}}{Q}^{2} + \frac{1}{2} < 0\), i.e. \(\eta < \frac{1+\sqrt{1-\frac{QL}{p_{k}^{2}}}}{L}\), the expectation of \(F_{k}\) keeps decreasing until \(\mathbb{E}( \|\nabla F_{k}(w_{k}^{t})\|^{2}) < \frac{(L+1)Q\sigma^{2}}{(2\eta-L\eta^{2})p_{k}^{2}+Q}\). This means that, given \(\eta < \frac{1+\sqrt{1-\frac{QL}{p_{k}^{2}}}}{L}\), client \(k\)'s local model \(w_{k}\) will at least reach a \(\epsilon\)\emph{-critical point} (i.e. \(\|\nabla F_{k}(w_{k})\| \leq \epsilon\)), with \(\epsilon = \sqrt{\frac{(L+1)Q\sigma^{2}}{(2\eta-L\eta^{2})p_{k}^{2}+Q}}\). \par
\hfill \par
Let \(w_{k}^{0}\) be client \(k\)'s initial model, then the time complexity for client \(k\) to reach \(w_{k}^{\epsilon}\) is \(O(\frac{w_{k}^{0}-w_{k}^{\epsilon}}{\epsilon\eta})\). It is worth mentioning that \(\eta < \frac{1+\sqrt{1-\frac{QL}{p_{k}^{2}}}}{L}\) also satisfies \(\eta < \frac{2}{L}\) as \(1+\sqrt{1-\frac{QL}{p_{k}^{2}}} < 2\). \par
\hfill \par

% Appendix will not appear in the proceedings of AAAI. 

\end{document}